\pdfoutput=1

\documentclass[11pt]{article}

\usepackage[preprint]{acl}
\usepackage{amsmath}
\usepackage{times}
\usepackage{latexsym}
\usepackage{multirow}
\usepackage{graphicx}
\usepackage{subcaption}
\usepackage[T1]{fontenc}

\usepackage[utf8]{inputenc}

\usepackage{microtype}

\usepackage{inconsolata}
\usepackage{float}

\usepackage{graphicx}

\title{Teach Me to Trick: Exploring Adversarial Transferability via Knowledge Distillation}

\author{
    Siddhartha Pradhan\\
    \And Shikshya Shiwakoti\\ 
    \And  Neha Bathuri\\
    }

\begin{document}
\maketitle

\section*{Abstract}
Adversarial examples pose a significant threat to deep neural networks, particularly in black-box settings where attacks rely on the transferability of perturbations across models. In this paper, we investigate whether \emph{Knowledge Distillation} (KD), specifically from multiple heterogeneous teacher models, can enhance the generation of transferable adversarial examples. We train a lightweight student model using two KD strategies—curriculum-based switching and joint optimization—using ResNet-50 and DenseNet-161 as teachers. The student is then used to generate adversarial examples using FG, FGS, and PGD attacks, which are evaluated against a black-box target (GoogLeNet). Our findings show that student models trained via multi-teacher KD achieve attack success rates comparable to ensemble-based baselines, while reducing adversarial generation time by up to 6×. Additionally, an ablation study reveals that lower temperature and inclusion of hard-label supervision significantly improve transferability. These results highlight the potential of KD as a tool not only for model compression, but also for strengthening black-box adversarial attacks.

\section{Introduction}

Despite the impressive performance of deep neural networks (DNNs) in image classification tasks, they remain highly vulnerable to adversarial examples—inputs perturbed with imperceptible noise that can mislead models into incorrect predictions \citep{goodfellow2015explaining, szegedy2013intriguing}. A particularly concerning aspect of these examples is their \emph{transferability}: adversarial inputs crafted for one model often succeed against others, including black-box targets \citep{papernot2016transferability, liu2017delving}.

In this paper, we explore whether \emph{Knowledge Distillation} (KD) \citep{Hinton2015}—a technique originally proposed to compress large models by transferring their knowledge into smaller models—can be leveraged to generate more transferable non-targeted adversarial examples. Specifically, we investigate whether training a student model via multi-teacher KD can produce adversarial inputs that generalize better in black-box settings. By distilling knowledge from heterogeneous teacher models (e.g., ResNet and DenseNet), the student may acquire a more diverse representation of decision boundaries, potentially enhancing the transferability of the adversarial examples it generates.

To test this hypothesis, we propose two distinct KD training strategies: (i) curriculum-based distillation, where the student is trained with different teachers across epochs, and (ii) joint optimization, where the student simultaneously aligns with multiple teacher outputs. We then evaluate the effectiveness of the student-generated adversarial examples under multiple attack methods: FG, FGS, and PGD, and measure their transferability against a black-box target model.

Our contributions are summarized as follows:
\begin{itemize}
    \item We propose a novel use of multi-teacher knowledge distillation to train student models for generating transferable adversarial examples.
    \item We introduce and compare two multi-teacher KD strategies: curriculum-based switching and joint optimization.
    \item We evaluate adversarial transferability using standard attack methods (FG, FGS, PGD) and demonstrate that student models trained via multi-teacher KD achieve comparable or superior performance to larger teacher ensembles, with significantly lower computational cost.
    \item We conduct an ablation study to analyze the impact of KD hyperparameters—temperature ($\tau$) and loss weight ($\alpha$)—on adversarial performance.
\end{itemize}

Our findings suggest that distilling knowledge from multiple, diverse teachers leads to student models that are not only lightweight and efficient, but also capable of generating adversarial examples with high black-box transferability.

\section{Background}
\subsection{Adversarial Examples}
Adversarial examples are inputs to machine learning models that have been intentionally perturbed in a way that causes the model to make incorrect predictions, despite the perturbations being imperceptible or negligible to humans. This vulnerability was first widely publicized by Szegedy et al. (2014) \cite{szegedy2014intriguing}, who demonstrated that adding carefully crafted noise to an image could drastically change a neural network’s output while the image appeared unchanged to the human eye.

Formally, given an input $x$, a trained model $f$ and ground truth label $y=f(x)$, an adversarial example $x^{\star}$ is crafted such that its distance to $x$ is under an assigned budget (i.e., $||x - x^{\star}||_p \leq B$) and $f(x^{\star}) \neq y$. In a \textit{non-targeted} attack, the adversarial input causes the model to make any prediction other than the ground truth label $f(x^{\star}) \neq y$. On the other hand, \textit{targeted attacks} are more specific: they cause the model to misclassify the $x^{\star}$ to a specific target class $y^{\star} = f(x^{\star})$ where $y^{\star} \neq y$.

\subsubsection{Generating Non-Targeted Adversarial Examples}
\textit{Fast Gradient Sign} (FGS), introduced by Goodfellow et al.~\cite{goodfellow2015explaining}, is a simple yet effective technique that perturbs the input in the direction of the sign of the gradient of the loss function with respect to the input. Given a model \( f \), an input \( x \), true label \( y \), and loss function \( \mathcal{L}(f(x), y) \), a non-targeted adversarial example $x^{\star} = x + \epsilon \cdot \text{sign}\left(\nabla_x \mathcal{L}(f(x), y)\right)$. Here, \( \epsilon \) is a small constant that controls the perturbation magnitude. FGS is an \( \ell_\infty \)-bounded, one-step attack that is computationally efficient and commonly used to establish a baseline for adversarial robustness.

Similarly, \textit{Fast Gradient} (FG) perturbs the input in the direction of the gradient, but constrains perturbations under the \( \ell_2 \) norm. The non-targeted adversarial example $x^{\star} = x + \epsilon \cdot \scriptstyle \frac{\nabla_x \mathcal{L}(f(x), y)}{\|\nabla_x \mathcal{L}(f(x), y)\|_2}$.

While both FGS and FG are one-step attacks, \textit{Projected Gradient Descent} (PGD), introduced by Madry et al.~\cite{madry2017towards}, is an iterative variant of FGS. It is widely considered as the most powerful first-order adversarial attack. PGD performs multiple small perturbation steps with magnitude \(\alpha\), and projects the adversarial example back to the \( \epsilon \)-ball around the original input after each update. The update rule for $x^{t+1} = \Pi_{\mathcal{B}_\epsilon(x)} \left( x^t + \alpha \cdot \text{sign}\left( \nabla_x \mathcal{L}(f(x^t), y) \right) \right)$. Here, \( \Pi_{\mathcal{B}_\epsilon(x)} \) denotes the projection operation to keep \( x^t \) within the \( \epsilon \)-ball centered at \( x \), and \( \alpha \) is the step size. PGD is widely used in adversarial training and robustness benchmarks due to its effectiveness and simplicity.

\subsubsection{Transferability of Adversarial Examples}

An adversarial example is considered transferable when it successfully misleads not only the model it was crafted for, but also other models—potentially with different architectures or trained on different datasets \citep{papernot2016transferability}. This phenomenon enables practical \textit{black-box attacks}, where attackers can train a substitute model and generate adversarial examples against it that transfer effectively to a target model without requiring access to the target’s parameters or training data \citep{liu2017delving,tramer2017space}.

To further enhance transferability in black-box settings, recent work has explored meta-learning-based approaches. For example, \citet{qin2019meta} proposed a meta-surrogate model that learns to optimize adversarial examples over a distribution of surrogate models, improving their generalization to unseen targets. Similarly, \citet{yin2020meta} introduced a generalizable meta-learning strategy that strengthens adversarial transfer across diverse architectures, thereby increasing the effectiveness of black-box attacks. These advances highlight the importance of improving model robustness in adversarial settings and deepen our understanding of transferable vulnerabilities in modern AI systems.

\subsection{Knowledge Distillation}
Knowledge distillation (KD) is a training method where a student model learns to approximate the output distribution of a teacher model that is generally much larger. In a typical setting, KD consists of two stages. First, a large teacher model is trained on diverse data sources to achieve high performance and generalization to unseen images. Then, a smaller student network is trained to approximate the teacher's internal representations and decision boundaries. Hence, it can be thought of as a compression method \cite{Gou_2021}. 

The soft targets from a high-capacity teacher model may effectively convey nuanced inter-class relationships \cite{Hinton2015}. In other words, this process ensures that the smaller student model produces an output probability distribution closely matching the teacher's. Additionally, the use of soft targets (probability distributions) instead of hard labels (true labels), acts as an implicit regularizer and aids in better generalization \citep{Hinton2015}. This process enables the compression of large, complex models into smaller, more efficient ones while retaining much of the teacher's knowledge, with comparable if not better performance \cite{Gou_2021}.

Formally, given a student network \(S\) and a pre-trained teacher network \(T\), the \textit{soft-target} loss is defined as the Kullback–Leibler (KL) divergence between the softmax outputs of the teacher \( J(x) \) and the student \(P(x)\):

\begin{equation}\label{loss-kl-eq}
    \mathcal{L}_{\text{soft}} = D_{\text{KL}}( J(x) \, || \, P(x))
\end{equation}

Additionally, it is typical for the raw softmax output of the teacher to be softened using a temperature parameter $\tau$, such that \( J_i = \frac{\exp(z_i / \tau)}{\sum_j \exp(z_j / \tau)} \), where $z_i$ is the logit for class $i$. Larger values of $\tau$ help student models place more importance on the relative probabilities assigned by the teachers to incorrect classes, often leading to better generalization \cite{Hinton2015, Gou_2021}.

Furthermore, since teacher outputs may occasionally be unreliable or misaligned with the ground truth, the student is further regularized using a supervised \textit{hard-label} loss. This is implemented as the cross-entropy between the one-hot encoded ground-truth label $1_{y}$ and the student's predicted distribution $P(x)$:

\begin{equation}\label{loss-ce-eq}
    \mathcal{L}_{\text{hard}} = - \sum 1_{y}\cdot\log P(x)
\end{equation}

The final training loss for the student combines both soft-label and hard-label components via a balancing hyperparameter \( \alpha \):

\begin{equation}\label{loss-total-eq}
    \mathcal{L_{KD}} = \alpha \cdot\mathcal{L}_{\text{hard}} + (1 - \alpha)\cdot\mathcal{L}_{\text{soft}}
\end{equation}

\subsubsection{Multiple Teachers}
Prior works have extended KD to use multiple teachers, allowing the student to mimic the output of an different teacher models by learning varied internal representations and decision boundaries \cite{Hinton2015, zhang2017deepmutuallearning, Gou_2021}. Notable and distinct methods for multi-teacher KD include: curriculum based learning, such as \citep{fukuda17_interspeech}, in which a teacher was selected from a pool of teachers at each iteration; and ensemble-based such as \citep{Hinton2015}, which used averaged logits from all teachers.


\section{Methodology}
In this section we describe our proposed method to investigate the effectiveness of Knowledge Distillation (KD) in generating adversarial examples to ultimately attack a black box model. We will be focusing on creating adversarial examples for image classification models and verify the transferability of the resulting perturbed images.

\subsection{Leveraging Teacher Models for Transferable Attacks}\label{sec:kd-setup}
Multiple teacher models can offer complementary strengths, enabling the student to generate adversarial examples that are more likely to succeed across different architectures and successfully transfer to a black box setting. However, challenges such as overfitting or diminishing returns may arise if the models are too similar. Hence, we utilize two heterogeneous architectures for teachers:
\begin{enumerate}
    \item ResNet-50: ResNet is a core architecture in modern computer vision tasks. It utilizes residual connections that allow the input to skip one or more layers and be added to the output of deeper layers. ResNet-50 contains 50 layers and approximately 23 million parameters.
    \item DenseNet-161: DenseNet connects each layer to every other layer within a block which strengthens feature propagation, improves feature reuse and fine-grained feature extraction. DenseNet-161 has 161 layers with approximately 26 million parameters.
\end{enumerate}

To test the transferability of the generated adversarial images in a black box setting we used the GoogLeNet architecture which stacks multiple types of convolutions (1x1, 3x3, 5x5) within the same layer. This allows the network to capture a wide variety of spatial features at different scales. This model is 22 layers deep and contains approximately 5.5 million parameters. 

Finally, to maintain consistency across different types of multi-teacher setups we selected ResNet-18 as the architecture for the student. This is a relatively small model that is 18 layers deep and consists of approximately 11 million parameters  

\begin{figure*}[h]
  \centering
  \includegraphics[width=0.7\linewidth]{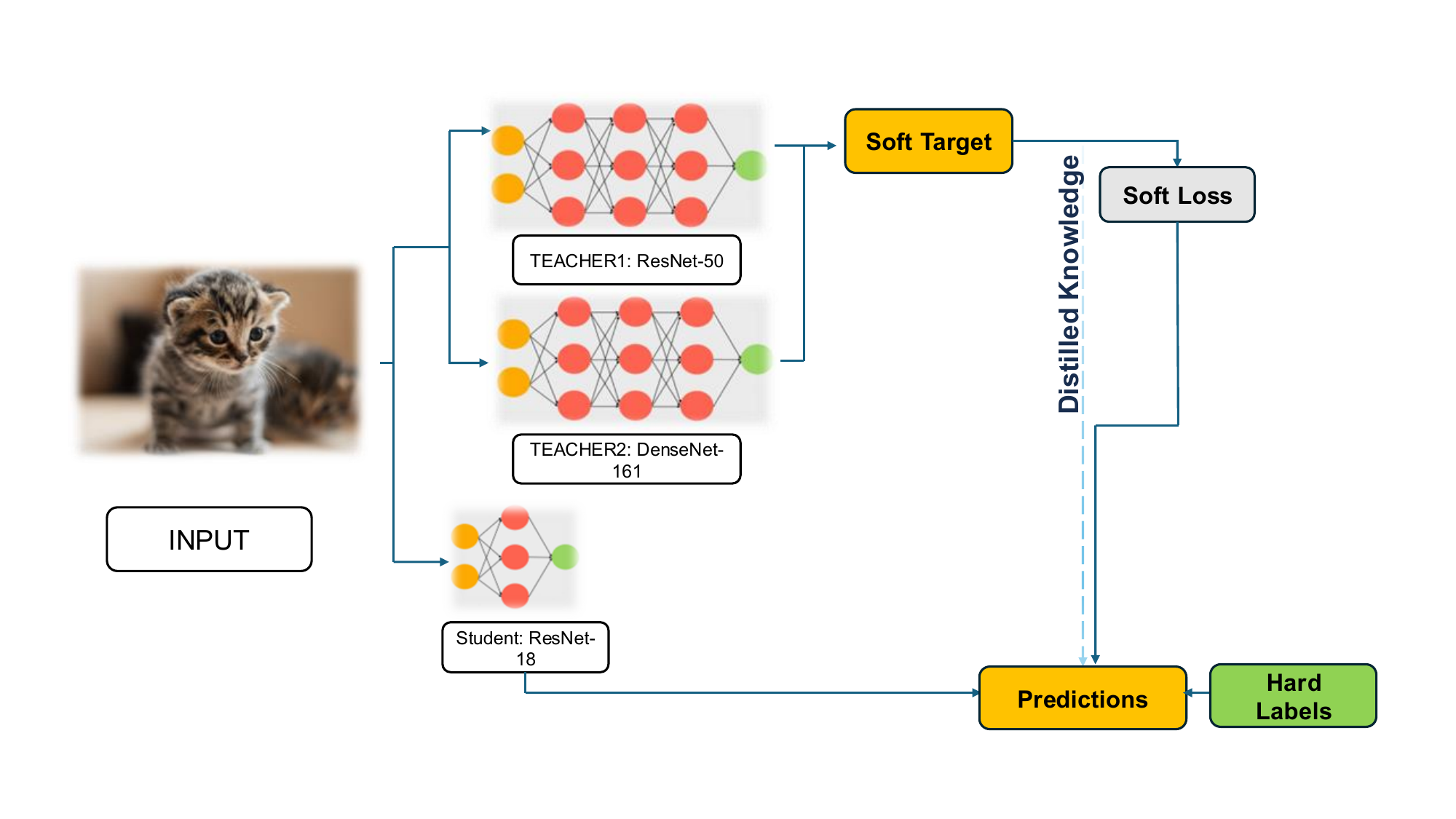}
  \caption {General Workflow for our proposed Knowledge Distillation Training Setup.}\label{fig:teacher-setup}
\end{figure*}

\subsubsection{Multi-Teacher Setup}
Figure~\ref{fig:teacher-setup} shows the general workflow of our proposed multi-teacher setup. The student's predictions and the soft targets extracted from the pre-trained teacher(s) are used to calculate $L_{soft}$ (Eq.~\ref{loss-kl-eq}). This transfers the decision boundaries of the teacher to the student. In addition, the student's predictions and ground-truth hard labels are used to compute the $L_{hard}$ (Eq.~\ref{loss-ce-eq}). The student is trained to optimized the combined distillation loss $L_{KD}$ (Eq.~\ref{loss-total-eq}). 

To examine the effects of multi-teacher knowledge distillation on adversarial transferability, we explore two distinct multi-teacher Knowledge Distillation setups, each differing in how the student network interacts (i.e. $L_{soft}$) with its teacher(s) during training.  

We explore the following two student types/setups:
\begin{itemize}
    \item \textbf{Type 1. Curricula based:} Similar to \citep{fukuda17_interspeech} we select an alternative teachers based on the epoch. However, switching every epoch could lead to unstable training. On the other hand, a long switching period could potentially lead to catastrophic forgetting between teacher changes, reducing the effectiveness of a multi-teacher setup. Hence, we change the teacher every 4 epochs, 0.04 of the maximum allocated epochs.
    \item \textbf{Type 2. Joint Optimization:} Unlike the ensemble based learning in \citep{Hinton2015} that averages the teacher logits before computing the soft-target loss \ref{loss-kl-eq}, we jointly compare the students output to all teachers. In our implementation we apply equal weight to each teacher. In other words, $L_{soft}$ is calculated for each teacher and averaged during each update step. As the student model learns to mimic the predictions of both teachers simultaneously, the output of the student model settles in the geometric center of all the teachers' predictions which may boost the transferability of the generated adversarial examples.
\end{itemize}

\subsection{Adversarial Attack Generation and Metrics}
In this work, we use three widely studied adversarial attack methods for generating adversarial examples using the student model: FGS, FG, and PGD. FG and FGS are efficient one-step attacks that perturb inputs using the student's gradient—FG uses the $\ell_2$ normalized gradient, while FGS uses its sign. Though fast, they often generate less transferable examples, particularly when the attacker's gradients and target's gradients are orthogonal \cite{liu2017delving}. PGD is an iterative variant of FGS that applies multiple small perturbations with projection, making it more effective and transferable, especially against defended models and cases where the gradients are orthogonal.

\subsubsection{Evaluation Metrics}
We use \textit{Attack Success Rate} (ASR) to evaluate the transferability and effectiveness of the attacks. ASR is the proportion of adversarial inputs that successfully cause misclassifications to the target model. ASR reflects the vulnerability of the model to the crafted adversarial samples. Next, to measure the practical feasibility of real-time or large-scale attacks we record the \textit{time taken} (in seconds) to generate an adversarial example using PGD. Finally, similar to \cite{liu2017delving}, we use Root Mean Square Deviation (RMSD) to measure the level of distortion between the original input image $x$ and its adversarial counterpart $x^{\star}$. It is computed as:
\[
\text{RMSD}(x, x^{\star}) = \sqrt{\frac{1}{N} \sum_{i=1}^{N} (x^{\star}_i - x_i)^2},
\]
where $x$ and $x^{\star}$ are the vectorized representations of the original and adversarial images respectively, $x_i$ and $x^{\star}_i$ denote the $i$-th pixel values (within the range $[0, 255]$), and $N$ is the total number of pixels. Lower RMSD indicates that the perturbation is more imperceptible to humans, thus better preserving visual similarity \cite{carlini2017towards}.

These metrics collectively help quantify the \textit{effectiveness}, \textit{efficiency}, and \textit{stealthiness} of adversarial attacks under a multi-teacher knowledge distillation setting.

\subsection{Experimental Setup}\label{sec:exp-setup}
\paragraph{Dataset.} We use CIFAR-10 as our dataset, which is comprised of 32$\times$32 color images across 10 object categories. The training set consists of 60{,}000 images, and the test set contains 10{,}000 images. Both teacher models and the black-box model are pretrained on CIFAR-10 prior to any student training or adversarial attack generation\footnote{Pretrained model weights can be accessed through \href{https://github.com/huyvnphan/PyTorch_CIFAR10/tree/master}{this} link.}. 

\paragraph{Training Configuration.} All experiments are conducted using a single NVIDIA RTX 3070-M GPU. We use a batch size of 256, optimized using Adam with weight-decay ($1\mathrm{e}{-6}$). Training is run for a maximum of 100 epochs, with early stopping applied based on validation loss plateau (patience of 10 epochs). Additionally, we use a learning rate scheduler with linear warmup for 30 epochs, which is subsequently decayed using a cosine schedule. We replicate the experiments with maximum $lr \in\{1\mathrm{e}{-2}, 1\mathrm{e}{-3}\}$. To ensure reproducibility, each setup is trained with 2 random seeds, and we report results from the models achieving the higher validation accuracy. The code for this project is publicly available \href{https://github.com/SiddharthaPradhan/KD-PGD.git}{here}. 

\paragraph{Knowledge Distillation Parameters.} Student networks are trained using a weighted combination of soft-target loss and hard-label cross-entropy loss, as defined in Section~\ref{sec:kd-setup}. To investigate the impact of distillation parameters, we tested two temperature settings, $\tau \in \{1, 5\}$, for softening teacher outputs, and two weighting coefficients, $\alpha \in \{0, 0.3\}$, to balance the hard-label and soft-label losses.

\paragraph{Attack Evaluation.} All adversarial attacks are generated on the test set. For PGD, the perturbation magnitude $\epsilon$, step size $\alpha$, and number of iterations are held constant across experiments. Similarly, for FG and FGS $\epsilon$ are held constant. More importantly, similar to \cite{liu2017delving}, for each attack we allocate a RMSD budget of $\approx 25$, ensuring a fair comparison across different attack types.

\subsection{Comparison Between Current Work and Initial Proposal}
We made several changes to the original experimental plan, driven by additional background research and experimentation:

\begin{enumerate}
    \item \textbf{Multiple Knowledge Distillation Schemes.} Instead of sticking to a single distillation strategy, we explored two distinct approaches: (i) curriculum-based—switching teachers at intervals, and (ii) joint optimization— simultaneous training on both teachers via a joint loss. This comparison allowed us to investigate whether structured exposure or joint knowledge distillation leads to better student generalization and attack transferability.

    \item \textbf{Teacher Ensemble for Transferability Baseline.} We moved beyond individual teacher comparisons by including attacks generated by an ensemble of both teachers for an additional baseline. This follows from prior work that shows ensembles improve attack success against black-box models~\cite{liu2017delving}.

    \item \textbf{Exclusion of Intermediate Representation Matching.} While methods like FitNets~\cite{romero2015fitnetshintsdeepnets} leverage internal feature maps, we chose not to match intermediate representations due to architectural mismatches between teachers (i.e., ResNet vs. DenseNet). We limited our scope to soft-target distillation to preserve architectural flexibility and isolate the effects of logit-based KD.

    \item \textbf{Ablation Study.} We conducted an ablation study across key hyperparameters: temperature $\tau$ and loss weighting $\alpha$. This analysis helps clarify the contribution of softened logits and the hard-vs-soft label balance in downstream adversarial transfer performance.
\end{enumerate}

\section{Results}

\subsection{Assessing Attack Transferability on the Black-Box Model}

For each model--including teachers, black-box and students--we generate adversarial examples using FGS, FG and PGD. We also include ensemble based attacks using both teachers, this was generated by averaging the teacher outputs with equal weights. We report the student models with the best transferability: trained with higher learning rate (i.e. $1\mathrm{e}{-2}$), $\alpha=0.3$ and $\tau=1$. Results for the student models for the experiment with lower learning rates have been provided in Appendix~\ref{app:exp-1}.

As described in Section~\ref{sec:exp-setup}, we chose the hyperparameters of the attack methods to maintain an RMSD of 25 $\pm1$. The attacks were generated on the test images ($N=10,000$) across all 10 classes and subsequently used to attack the black-box GoogLeNet model. Additionally, we limit the number of iterations to generate PGD attacks to 10 with a batch size of 150 and recorded the total time taken for the entire test set. We present these results in Table~\ref{tab:asr_comparison}.

For the baseline models, the teacher-ensemble based attacks outperformed individual teachers in attack transferability for the PGD based attack (\textbf{96\%} vs. 93\% for ResNet-50, and 91\% for DenseNet-151). On the other, hand our proposed student model with curriculum training performs comparably to the Ensemble baseline (\textbf{95\%} for PGD). Most notably, the time taken to generate 10,000 PGD attacks for our method took only \textbf{33} seconds. In comparison to the ensemble based PGD which took 200 seconds, our student model is 6 times faster for PGD attack generation with comparable adversarial transferability.

\begin{table*}[h]
\centering
\begin{tabular}{|c|l|c|c|c|c|c|}
\hline
\textbf{Type} & \textbf{Attacker Model} & \textbf{RMSD} & \textbf{FG} & \textbf{FGS} & \textbf{PGD} & \textbf{PGD Time (s)} \\
\hline
\textbf{Self-Attack} 
  & GoogLeNet (Blackbox)               & 24.48 & 0.80 & 0.88 & 1.00 & 176.47 \\
\hline
\multirow{3}{*}{\textbf{Baselines}} 
  & ResNet-50 (Teacher 1)              & 24.49 & 0.69 & 0.78 & 0.93 & 69.60 \\
  & DenseNet-151 (Teacher 2)           & 24.48 & 0.67 & 0.76 & 0.91 & 139.58 \\
  & Ensemble (ResNet-50 \& DenseNet-151) & 24.48 & 0.69 & 0.77 & \textbf{0.96} & 201.21 \\
\hline
\multirow{2}{*}{\textbf{Students}} 
  & Curriculum Trained (Type 1)        & 24.56 & \textbf{0.78} & \textbf{0.86} & \textbf{0.95} & \textbf{33.01} \\
  & Jointly Trained (Type 2)           & 24.55 & 0.73 & 0.83 & 0.93 & \textbf{32.03} \\
\hline
\end{tabular}
\caption{RMSD, ASRs against Blackbox, and PGD attack generation runtime across models.}
\label{tab:asr_comparison}
\end{table*}
 
\subsection{Ablation Study: Effects of KD Parameters For Attack Transferability}
We conduct an ablation study to assess how varying the temperature parameter ($\tau$) and mixing weight ($\alpha$) for $L_{soft}$ (Eq.~\ref{loss-ce-eq}) affects adversarial transferability. The results (Table~\ref{tab:kd_ablation}) demonstrate two clear trends.

\begin{table*}[h!]
\centering
\begin{tabular}{|c|c|c|c|c|c|c|c|}
\hline
\textbf{Student Type} & \textbf{KD Params} & \textbf{RMSD} & \textbf{FG} & \textbf{FGS} & \textbf{PGD} & \textbf{Acc/Test Set} & \textbf{PGD Time (s)} \\
\hline
\multirow{4}{*}{\textbf{\shortstack{\textbf{Type 1:}\\\textbf{Curriculum}}}} 
  & $\alpha=0$, $\tau=1$   & 24.56 & 0.74 & 0.82 & 0.94 & 0.88 & 33.01 \\
  & $\alpha=0$, $\tau=5$   & 24.56 & 0.71 & 0.80 & 0.85 & 0.88 & 32.45 \\
  & $\alpha=0.3$, $\tau=1$ & 24.56 & 0.78 & 0.86 & 0.95 & 0.87 & 33.97 \\
  & $\alpha=0.3$, $\tau=5$ & 24.56 & 0.68 & 0.78 & 0.85 & 0.88 & 33.21 \\
\hline
\multirow{4}{*}{\textbf{\shortstack{\textbf{Type 2:}\\\textbf{Joint}}}} 
  & $\alpha=0$, $\tau=1$   & 24.56 & 0.76 & 0.84 & 0.93 & 0.90 & 32.03 \\
  & $\alpha=0$, $\tau=5$   & 24.55 & 0.71 & 0.79 & 0.90 & 0.91 & 33.84 \\
  & $\alpha=0.3$, $\tau=1$ & 24.55 & 0.73 & 0.83 & 0.93 & 0.88 & 32.50 \\
  & $\alpha=0.3$, $\tau=5$ & 24.55 & 0.71 & 0.81 & 0.88 & 0.89 & 33.09 \\
\hline
\end{tabular}
\caption{Comparison of ASR against Blackbox model across different KD configurations for Student Types 1 and 2.}
\label{tab:kd_ablation}
\end{table*}

First, increasing the temperature from $\tau=1$ to $5$ consistently leads to worse attack performance across all metrics (FG, FGS, PGD), indicating reduced transferability of adversarial examples. The overly softened outputs from the teacher may have obscured decision boundaries necessary for effective gradient alignment.

Second, incorporating a hard-label component by setting $\alpha=0.3$ (as opposed to $\alpha=0$) improves transferability in most settings, particularly under lower temperature. The combination $\alpha=0.3$, $\tau=1$ consistently yields strong ASRs across both student types. This suggests that retaining some supervision from the ground-truth labels further regularizes the parameters, enhancing the students' ability to generate transerable attacks.

\subsection{Visualizing The Result of Different Attack Types}

\begin{figure*}[h!]
    \centering
    \includegraphics[width=0.6\linewidth]{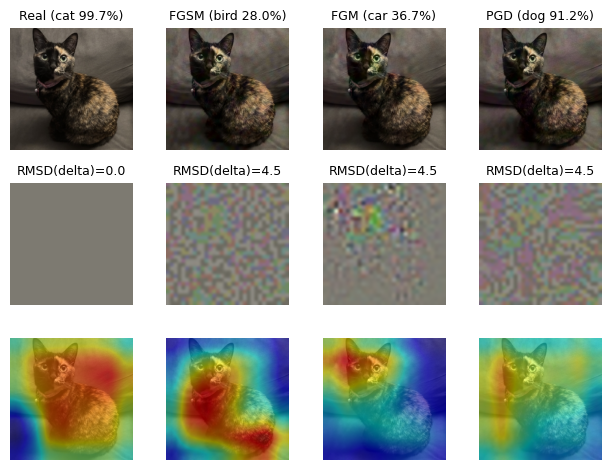}
    \caption{Adversarial examples and Grad-CAM attributions. Top: predictions and confidences for original and adversarial images. Middle: scaled perturbations with RMSD. Bottom: Grad-CAM maps showing attention shifts under FGSM, FGM, and PGD.}
    \label{fig:mochi-adv-example}
\end{figure*}

Figure~\ref{fig:mochi-adv-example} demonstrates the impact of adversarial attacks FGSM, FGM, and PGD on a CIFAR-like input\footnote{The image was downscaled to 32x32 for compatibility. However, the resulting perturbations and Saliency Maps are upscaled and added to the original image for visual clarity. The image is Sid's cat named ``Mochi''.}. Despite minimal RMSD differences ($\simeq 4.5$), adversarial perturbations cause large shifts in model predictions.

The second row shows the visualized perturbations scaled for interpretability. Although visually imperceptible, these perturbations are sufficient to significantly alter predictions.

The third row presents Grad-CAM saliency overlays. For the clean image, the blackbox model focuses on the outline of the cat, whereas adversarial variants resulted in degraded and dispersed attention. This saliency map highlights the distortion in the model's internal reasoning that eventually leads to the misclassifications.

\section{Discussion}
This work explores the potential of knowledge distillation (KD) for adversarial transferability, specifically in black-box attack settings. By distilling knowledge from multiple heterogeneous teacher models into a single student model and using this student to generate adversarial examples, we assessed whether KD can improve the efficiency and effectiveness of first-order adversarial attacks. Our findings offer several important insights.

Our method demonstrates performance that is comparable to ensemble-based approaches under PGD attacks, while being 6 times faster. This efficiency presents a practical advantage: faster adversarial example generation makes the method well-suited for integration into adversarial training pipelines, where computational cost is a limiting factor. By leveraging this speedup, it becomes feasible to incorporate high-quality adversarial examples into the training process without incurring the significant overhead associated with ensemble-based methods. 

\subsection{Exploring the geometric properties of the learned decision boundaries}
We observe superior adversarial transferability under single-step attacks such as FG and FGS. This suggests that the student model's initial gradient direction more closely aligns with that of the black-box model, enabling better alignment in early adversarial perturbations. However, this advantage diminishes under multi-step attacks like PGD, where the student model exhibits weaker transferability. This discrepancy implies that while the student may capture coarse decision boundaries that suffice for initial alignment, it likely fails to internalize more nuanced and global boundary structures needed for stronger generalization. 

\begin{figure}[ht!]
    \centering
    \begin{subfigure}[b]{0.47\textwidth}
        \centering
        \includegraphics[width=\textwidth]{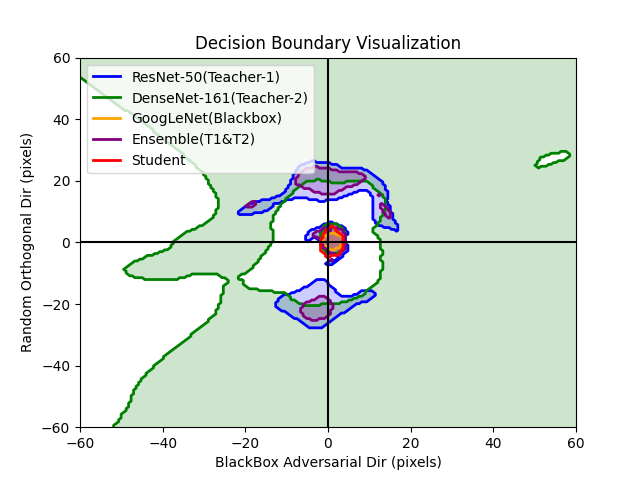}
        \caption{Zoomed out decision boundary with perturbations up to 50 pixels.}
        \label{fig:zoom_out}
    \end{subfigure}
    \begin{subfigure}[b]{0.47\textwidth}
        \centering
        \includegraphics[width=\textwidth]{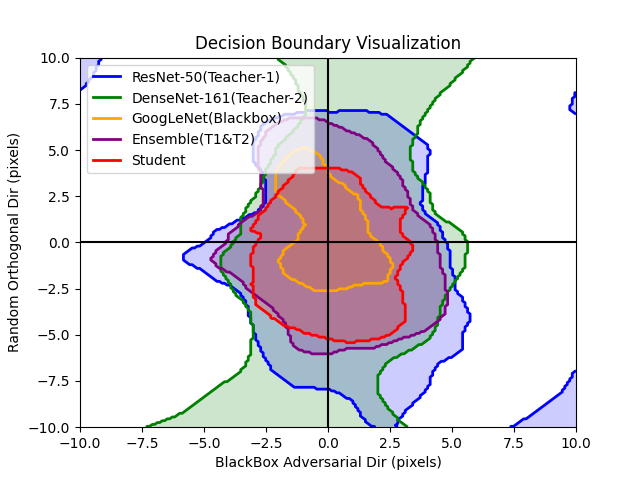}
        \caption{Zoomed in decision boundary with perturbations up to 6 pixels.}
        \label{fig:zoom_in}
    \end{subfigure}
    \caption{Decision Boundaries for models using in this study. The X-Axis represents the normalized gradient direction for the Black Box model. The Y-Axis is a random orthogonal direction. The real image ($x$) is from Figure~\ref{fig:mochi-adv-example} and is located at the origin.}
    \label{fig:decision-boundary}
\end{figure}

To explore this further, we study the decision boundaries of models included in the study. We provide this in Figure~\ref{fig:decision-boundary}. Figure~\ref{fig:zoom_out} shows a wide field of view ($\pm50$ pixels), while Figure~\ref{fig:zoom_in} zooms into the immediate neighborhood ($\pm6$ pixels) around the original input. Additionally shaded regions coincide to regions where the respective model correctly predicted the correct class (in this case a cat).

Globally (Figure~\ref{fig:zoom_out}), the student’s boundary (red) is a tight, almost circular region very near the origin, whereas the individual teachers (green and blue) and the ensemble's boundary (purple) carve out large, non-convex regions. The student’s failure to cover these wide regions explains why, under PGD (which traverses farther from the origin), its attack success falls off relative to an ensemble: the student simply never learned those more remote decision boundaries.

Locally (Figure~\ref{fig:zoom_in}), however, the student’s red contour nearly coincides with the black-box (orange) and ensemble (purple) contours within a few pixels of the origin.  This local boundary alignment may explain the strong FG and FGS performance: single-step attacks rely only on the initial gradient direction, which the student has faithfully captured. 

\section{Limitations}
While our logit‐level distillation yields fast, locally accurate adversarial attacks, it does not capture the global decision geometry of teachers or ensembles—limiting multi‐step (e.g. PGD) transferability. Additionally, our study uses just two teachers and fixed distillation schedules; richer teacher pools, adaptive scheduling, or intermediate‐feature matching could further improve transferability. Moreover, we validate only on CIFAR-10 and a single GoogLeNet black-box; results may differ on higher-resolution benchmarks or other target architectures.   

\bibliography{custom}

\begin{thebibliography}{15}
\providecommand{\natexlab}[1]{#1}

\bibitem[{Carlini and Wagner(2017)}]{carlini2017towards}
Nicholas Carlini and David Wagner. 2017.
\newblock \href {https://arxiv.org/abs/1608.04644} {Towards evaluating the robustness of neural networks}.
\newblock \emph{IEEE Symposium on Security and Privacy (SP)}.

\bibitem[{Fukuda et~al.(2017)Fukuda, Suzuki, Kurata, Thomas, Cui, and Ramabhadran}]{fukuda17_interspeech}
Takashi Fukuda, Masayuki Suzuki, Gakuto Kurata, Samuel Thomas, Jia Cui, and Bhuvana Ramabhadran. 2017.
\newblock \href {https://doi.org/10.21437/Interspeech.2017-614} {Efficient knowledge distillation from an ensemble of teachers}.
\newblock In \emph{Interspeech 2017}, pages 3697--3701.

\bibitem[{Goodfellow et~al.(2015)Goodfellow, Shlens, and Szegedy}]{goodfellow2015explaining}
Ian~J Goodfellow, Jonathon Shlens, and Christian Szegedy. 2015.
\newblock Explaining and harnessing adversarial examples.
\newblock \emph{arXiv preprint arXiv:1412.6572}.

\bibitem[{Gou et~al.(2021)Gou, Yu, Maybank, and Tao}]{Gou_2021}
Jianping Gou, Baosheng Yu, Stephen~J. Maybank, and Dacheng Tao. 2021.
\newblock \href {https://doi.org/10.1007/s11263-021-01453-z} {Knowledge distillation: A survey}.
\newblock \emph{International Journal of Computer Vision}, 129(6):1789–1819.

\bibitem[{Hinton et~al.(2015)Hinton, Vinyals, and Dean}]{Hinton2015}
Geoffrey Hinton, Oriol Vinyals, and Jeff Dean. 2015.
\newblock \href {https://arxiv.org/abs/1503.02531} {Distilling the knowledge in a neural network}.
\newblock \emph{arXiv preprint arXiv:1503.02531}.

\bibitem[{Liu et~al.(2017)Liu, Chen, Liu, and Song}]{liu2017delving}
Yanpei Liu, Xinyun Chen, Chang Liu, and Dawn Song. 2017.
\newblock \href {https://arxiv.org/abs/1611.02770} {Delving into transferable adversarial examples and black-box attacks}.
\newblock In \emph{Proceedings of the 5th International Conference on Learning Representations (ICLR)}.

\bibitem[{Madry et~al.(2018)Madry, Makelov, Schmidt, Tsipras, and Vladu}]{madry2017towards}
Aleksander Madry, Aleksandar Makelov, Ludwig Schmidt, Dimitris Tsipras, and Adrian Vladu. 2018.
\newblock Towards deep learning models resistant to adversarial attacks.
\newblock In \emph{International Conference on Learning Representations (ICLR)}.

\bibitem[{Papernot et~al.(2016)Papernot, McDaniel, Goodfellow, Jha, Celik, and Swami}]{papernot2016transferability}
Nicolas Papernot, Patrick McDaniel, Ian Goodfellow, Somesh Jha, Z~Berkay Celik, and Ananthram Swami. 2016.
\newblock Transferability in machine learning: from phenomena to black-box attacks using adversarial samples.
\newblock In \emph{arXiv preprint arXiv:1605.07277}.

\bibitem[{Qin et~al.(2019)Qin, Xiong, Chen, Liu, and Song}]{qin2019meta}
Yunxiao Qin, Yuanhao Xiong, Xinyun Chen, Chang Liu, and Dawn Song. 2019.
\newblock Training a meta-surrogate model for transfer attacks.
\newblock In \emph{Proceedings of the 25th ACM SIGKDD International Conference on Knowledge Discovery \& Data Mining}, pages 1298--1306.

\bibitem[{Romero et~al.(2015)Romero, Ballas, Kahou, Chassang, Gatta, and Bengio}]{romero2015fitnetshintsdeepnets}
Adriana Romero, Nicolas Ballas, Samira~Ebrahimi Kahou, Antoine Chassang, Carlo Gatta, and Yoshua Bengio. 2015.
\newblock \href {https://arxiv.org/abs/1412.6550} {Fitnets: Hints for thin deep nets}.
\newblock \emph{Preprint}, arXiv:1412.6550.

\bibitem[{Szegedy et~al.(2013)Szegedy, Zaremba, Sutskever, Bruna, Erhan, Goodfellow, and Fergus}]{szegedy2013intriguing}
Christian Szegedy, Wojciech Zaremba, Ilya Sutskever, Joan Bruna, Dumitru Erhan, Ian Goodfellow, and Rob Fergus. 2013.
\newblock Intriguing properties of neural networks.
\newblock \emph{arXiv preprint arXiv:1312.6199}.

\bibitem[{Szegedy et~al.(2014)Szegedy, Zaremba, Sutskever, Bruna, Erhan, Goodfellow, and Fergus}]{szegedy2014intriguing}
Christian Szegedy, Wojciech Zaremba, Ilya Sutskever, Joan Bruna, Dumitru Erhan, Ian Goodfellow, and Rob Fergus. 2014.
\newblock Intriguing properties of neural networks.
\newblock \emph{arXiv preprint arXiv:1312.6199}.

\bibitem[{Tram{\`e}r et~al.(2017)Tram{\`e}r, Papernot, Goodfellow, Boneh, and McDaniel}]{tramer2017space}
Florian Tram{\`e}r, Nicolas Papernot, Ian Goodfellow, Dan Boneh, and Patrick McDaniel. 2017.
\newblock Space of transferable adversarial examples.
\newblock In \emph{arXiv preprint arXiv:1704.03453}.

\bibitem[{Yin et~al.(2020)Yin, Zhang, Li, Zhang, Tay, and Cheung}]{yin2020meta}
Fei Yin, Yong Zhang, Qianru Li, Yinghui Zhang, Yi~Tay, and Ngai-Man Cheung. 2020.
\newblock Generalizable black-box adversarial attack with meta learning.
\newblock In \emph{European Conference on Computer Vision (ECCV)}, pages 685--702. Springer.

\bibitem[{Zhang et~al.(2017)Zhang, Xiang, Hospedales, and Lu}]{zhang2017deepmutuallearning}
Ying Zhang, Tao Xiang, Timothy~M. Hospedales, and Huchuan Lu. 2017.
\newblock \href {https://arxiv.org/abs/1706.00384} {Deep mutual learning}.
\newblock \emph{Preprint}, arXiv:1706.00384.

\end{thebibliography}

\clearpage
\appendix
\onecolumn
\section{Experimental Results for Student Model trained with lower learning rate}\label{app:exp-1}
\begin{table*}[h!]
\small
\centering
\begin{tabular}{|c|c|c|c|c|c|c|c|}
\hline
\textbf{Student Type} & \textbf{KD Params} & \textbf{RMSD} & \textbf{FG} & \textbf{FGS} & \textbf{PGD} & \textbf{Acc/Test Set} & \textbf{PGD Time (Sec)} \\
\hline
\multirow{4}{*}{\textbf{\shortstack{\textbf{Type 1:}\\\textbf{Curriculum}}}} 
& $a=0, t=1$   & 24.48 & 0.79 & 0.85 & 0.90 & 0.90 & 33.01 \\
& $a=0, t=5$   & 24.48 & 0.67 & 0.76 & 0.83 & 0.90 & 32.45 \\
& $a=0.3, t=1$ & 24.48 & 0.77 & 0.84 & 0.88 & 0.90 & 33.97 \\
& $a=0.3, t=5$ & 24.48 & 0.66 & 0.75 & 0.85 & 0.90 & 33.21 \\
\hline
\multirow{4}{*}{\shortstack{\textbf{Type 2:}\\\textbf{Joint}}} 
& $a=0, t=1$   & 24.48 & 0.74 & 0.82 & 0.90 & 0.91 & 32.03 \\
& $a=0, t=5$   & 24.48 & 0.68 & 0.78 & 0.89 & 0.91 & 33.84 \\
& $a=0.3, t=1$ & 24.48 & 0.74 & 0.82 & 0.88 & 0.91 & 32.50 \\
& $a=0.3, t=5$ & 24.48 & 0.70 & 0.78 & 0.86 & 0.90 & 33.09 \\
\hline
\end{tabular}
\caption{Comparison of ASR against Blackbox model across different KD configurations for Student Types 1 and 2. Maximum $lr=1e-3$}
\label{tab:kd_performance}
\end{table*}

\end{document}